\documentclass[11pt,letterpaper]{article}
\usepackage[margin=1in]{geometry}
\usepackage[T1]{fontenc}
\usepackage[utf8]{inputenc}
\usepackage{tgtermes}
\usepackage{latexsym}
\usepackage{inconsolata}
\usepackage{microtype}
\usepackage{graphicx}
\usepackage{booktabs}
\usepackage{caption}
\usepackage{placeins}
\usepackage{needspace}
\usepackage{array}
\usepackage{amsmath,amssymb}
\usepackage[authoryear,round]{natbib}
\usepackage[hidelinks]{hyperref}
\usepackage{url}
\newsavebox{\papertablebox}

\newcommand{\experimentfigure}[2]{\includegraphics[width=#1]{#2}}
\title{Neyshekar: An Open Persian Read-Speech Corpus for Automatic Speech Recognition}
\author{\begin{tabular}{@{}c@{}}
{\normalsize Ahmad Amirivojdan, Farzad Nadiri, Abolfazl Alizadeh, Shaghayegh Yaraghi} \\
{\small Shekar AI}
\end{tabular}}
\date{}
\hypersetup{pdftitle={Neyshekar: An Open Persian Read-Speech Corpus for Automatic Speech Recognition},pdfauthor={Ahmad Amirivojdan, Farzad Nadiri, Abolfazl Alizadeh, Shaghayegh Yaraghi}}

\begin{document}
\maketitle

\begin{abstract}
Neyshekar is presented as an open Persian read-speech corpus designed for coverage of both formal and informal language, named entities, and longer utterances. In version 6, 62,279 validated recordings totalling 99.02 hours are provided from 190 contributors, with 34,541 distinct recorded prompts. The prompt pool was assembled from human-written material, contextualised homographs, and reviewed language-model-generated text. Text entries were normalised with the shekar library, which supports both formal and informal Persian, and every submitted recording was reviewed against a common validation rubric. About 24\% of released clips are classified as informal by an automatic classifier; these register labels are not human-validated. Item-level rater labels are provided for reproducible agreement estimation, opaque per-clip contributor identifiers make the speaker-disjoint partitioning auditable and support contributor-clustered uncertainty estimates, and a text-disjoint test subset is included for evaluation beyond previously seen prompts. Per-contributor recording load and reference-free signal quality are characterised for every released clip. Corpus characteristics are compared with Persian Common Voice under shared processing. Utility is assessed through two ASR architectures, three optimisation seeds, WER and CER, and independent evaluation on the public PSRB sample. Against duration-matched Common Voice training at approximately 32 hours, in-domain WER is reduced by 9.5 points for Whisper and 11.6 points for XLS-R, and by approximately eight points for both architectures on the independent PSRB sample. Transfer and mixture benefits are not consistently observed across architectures and training budgets. The corpus is released under CC0; code and data are made available through the project repository at \url{https://github.com/amirivojdan/neyshekar}.
\end{abstract}

\FloatBarrier
\section{Introduction}
\label{sec:introduction}

Persian is written in the Perso-Arabic script and is spoken by more than 100 million people, principally in Iran and Afghanistan~\citep{psrb2025}. Persistent ASR errors are nevertheless reported for informal speech, regional accents, and named entities~\citep{psrb2025}. Of these, informal register and named-entity coverage are addressed by the corpus presented here; regional accent is not, because accent is not recorded in the contributor metadata. Formal and informal Persian are distinguished by phonology and morphosyntax, and their separation has been discussed as diglossia~\citep{ferguson1959diglossia,mahmoodi2018diglossia}. Pronunciation ambiguity is compounded by the omission of short vowels in Persian orthography~\citep{fetrat2025g2p}. Broader register and lexical coverage is therefore motivated alongside increased audio volume.

Existing Persian resources are distinguished by recording style and intended use (Table~\ref{tab:related-corpora}). FARSDAT, MirasVoice, ShEMO, and DeepMine were developed for speech, speaker, or emotion recognition~\citep{bijankhan1994farsdat,vaheb2018mirasvoice,nezami2019shemo,zeinali2019deepmine}; narration has been used for TTS corpora~\citep{adibian2024deepmine,fetrat2025manatts,kalahroodi2025parsvoice}, while read speech has been collected through FLEURS and Common Voice~\citep{conneau2022fleurs,ardila2020commonvoice}. Spontaneous Persian Speech (SPS), comprising 694 minutes from 65 speakers, was introduced for Whisper adaptation~\citep{namdarzadeh2026sps}. Informal wording in read prompts should be distinguished from spontaneous production, and corpus quality cannot be inferred from size alone.

ASR degradation under domain mismatch has been reduced with target-domain audio~\citep{likhomanenko2021rethinking,hsu2021robust}. Condition-specific evaluation has also been reported for Urdu, Brazilian Portuguese, African-accented English, and Western Sierra Puebla Nahuatl~\citep{arif-etal-2025-wer,evaldo-leal-etal-2025-mupe,sanni-etal-2025-afrispeech,pugh-etal-2025-ihquin}. For Persian, varied speech sources and acoustic conditions are represented in the public PSRB sample~\citep{psrb2025}, through which evaluation beyond crowdsourced read speech is enabled.

\begin{table}[htbp]
\caption{Selected Persian speech corpora related to Neyshekar. Values are taken from published corpus descriptions, except for Common Voice and Neyshekar, for which values were computed from the analysed releases.}
\label{tab:related-corpora}
\centering
\small
\setlength{\tabcolsep}{5pt}
\renewcommand{\arraystretch}{1.1}
\begin{tabular}{@{}>{\raggedright\arraybackslash}p{3.3cm}p{4.0cm}rr@{}}
\toprule
Corpus & Speech and source & Hours & Speakers \\
\midrule
FARSDAT & Read; 10 dialect regions & -- & 300 \\
MirasVoice & Read + spontaneous Fa--En & $>$33 & 50 \\
ShEMO & Semi-natural emotional & 3.4 & 87 \\
DeepMine & Fa--En recordings & $>$480 & $>$1.8k \\
ManaTTS & Single-speaker narration & 86 & 1 \\
ParsVoice & Aligned audiobook segs. & 2{,}200 & 1{,}815 \\
FLEURS & Multilingual read speech & $\sim$12 & -- \\
Common Voice 26.0 & Crowdsourced prompts & 373.2 & 4{,}338 \\
Neyshekar v6 & Crowdsourced prompts & 99.0 & 190 \\
\bottomrule
\end{tabular}
\end{table}

Neyshekar is presented as a CC0 Persian read-speech corpus designed for longer utterances, both formal and informal language, and curated named-entity coverage. To date, 62{,}279 recordings totalling 99.02 hours have been validated. Version 6 is analysed throughout, and the label \emph{Neyshekar v6} is used where release-specific statistics are reported. The Common Voice collection approach was adapted through controlled prompt construction and review by trained validators. Collection procedures and annotation agreement are documented in Section~\ref{sec:construction}; corpus characteristics are compared with Persian Common Voice under identical processing in Section~\ref{sec:cv-comparison}.

Corpus utility is assessed with two ASR architectures and three optimisation seeds. Training duration and update budgets are controlled, and WER and CER are reported for the source corpus, text-disjoint prompts, Common Voice, and PSRB. Mixture and scaling experiments are included to distinguish corpus effects from additional training resources, with paired uncertainty estimates provided for the principal comparisons.

\section{Dataset Construction}
\label{sec:construction}

\FloatBarrier
\subsection{Prompt Design and Text Normalisation}
\begin{figure}[htbp]
\centering
\experimentfigure{0.85\columnwidth}{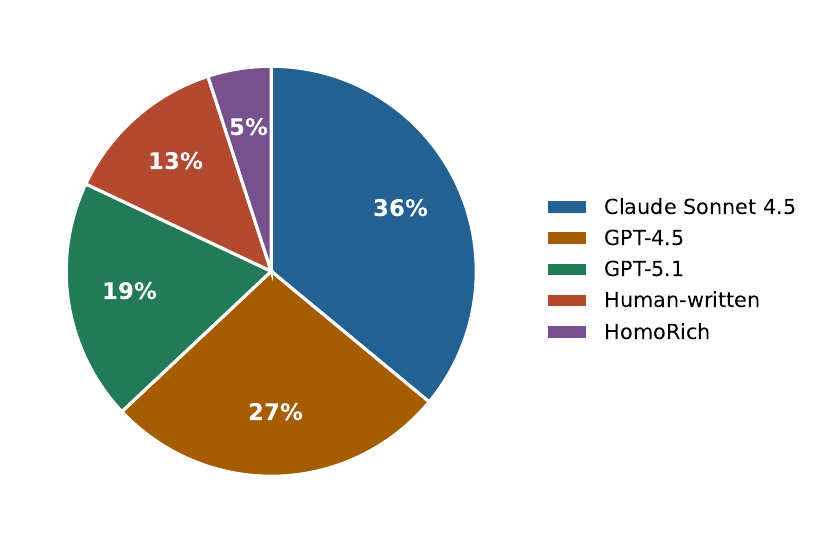}
\caption{Approximate prompt-pool composition reported by the authors. Shares refer to prompt sources, not recording counts or audio duration; generated sources together account for 82\% of the pool.}
\label{fig:prompt-sources}
\end{figure}

The prompt pool was assembled from three source groups for register coverage, named-entity density, and pronunciation ambiguity (Figure~\ref{fig:prompt-sources}). Approximately 5\% of prompts were drawn from the manually written portion of HomoRich, a CC0 Persian homograph corpus~\citep{fetrat2025g2p} in which homographs are placed in disambiguating contexts. Another 13\% were entered directly as short human-written everyday phrases. The remaining 82\% were generated with Claude Sonnet 4.5 (36\% of the full pool), GPT-4.5 (27\%), and GPT-5.1 (19\%)~\citep{anthropic2025sonnet45,openai2025gpt45,openai2025gpt51}, using instructions for varied topics, text types, and registers. In a subset, real public figures and places were explicitly requested: footballers, singers, professors, politicians, historic palaces, streets, monuments, and institutions. Every generated mention was subjected to sentence-level review before inclusion.

All text entries were normalised with the \texttt{shekar} library~\citep{shekar2025joss}. Both formal and informal Persian are supported by its normalisation pipeline, which was essential for consistent orthographic processing across the two registers. The entries were then sentence-segmented with \texttt{shekar} and reviewed sentence by sentence by an administrator. Candidates were accepted only when complete, fluent, correctly segmented, and consistent with the stated register. Linguistic quality was checked, and factual claims in model-generated text were verified. Naturalness equivalence to human-written prompts was not measured, however, and representativeness of everyday colloquial speech is not established by this review. In total, 36{,}554 sentences were admitted to the pool; 34{,}541 were recorded at least once and included in the release.

\FloatBarrier
\subsection{Speech Collection}
Recordings were collected through a purpose-built web application with a standard front end and API; audio was captured client-side and stored for validation. After account creation, one prompt was displayed at a time, with replay, re-record, submit, and skip options. Personal devices were used without device restrictions or a signal-to-noise threshold. Written Persian instructions aligned with the validation rubric were displayed before recording. Each prompt was required to be read exactly, completely, clearly, and fluently, without paraphrase or register substitution, in a quiet setting at a fixed microphone distance. Re-recording was required after any restart, hesitation, or correction, and replay before submission. In total, 69{,}744 recordings were submitted by 198 registered contributors against the 36{,}554-prompt pool.

\FloatBarrier
\subsection{Annotation and Quality Control}
All submitted recordings were reviewed by a principal validator and labelled as accepted or rejected under a six-criterion Persian rubric: added, omitted, replaced, or improvised words; restarts, repetitions, hesitations, or mid-speech corrections; incomplete sentences; pronunciation errors that changed a word; formal/informal substitution; or disruptive background noise. Recordings were rejected if any criterion was violated. Of 69{,}744 submissions, 62{,}279 (89.3\%) were accepted.

Three additional trained annotators were included only in a reliability study; release labels were determined solely by the principal validator. After training on a discarded 50-recording sample, a new random sample of 300 recordings was independently labelled by all four annotators (Table~\ref{tab:agreement}). Fleiss' multi-rater $\kappa$~\citep{fleiss1971kappa} and Gwet's AC1~\citep{gwet2008ac1} are reported as primary measures because four labels were available for each recording. Under the descriptive scale of \citet{landiskoch1977agreement}, $\kappa$ is classified as ``substantial'' and AC1 as ``almost perfect.'' These estimates are reproduced from item-level labels, with 20,000 bootstrap resamples of complete four-rater items. Acceptance and rejection agreement are estimated at 95.6\% and 72.3\%, respectively.

\begin{table}[htbp]
\centering
\caption{Inter-annotator reliability on the shared sample of 300 recordings (four independent annotators; 95\% CIs).}
\label{tab:agreement}
\normalsize
\setlength{\tabcolsep}{6pt}
\renewcommand{\arraystretch}{1.1}
\begin{tabular}{lr}
\toprule
Measure & Value \\
\midrule
Independent labels & 1,200 \\
Accept-label share & 86.3\% \\
Fleiss' $\kappa$ (95\% CI) & 0.679 [0.591, 0.757] \\
Gwet's AC1 (95\% CI) & 0.900 [0.866, 0.930] \\
Overall raw agreement & 92.4\% \\
Agreement on accept & 95.6\% \\
Agreement on reject & 72.3\% \\
\bottomrule
\end{tabular}
\end{table}

\FloatBarrier
\subsection{Data Preprocessing and Partitioning}
\label{sec:partitioning}
Accepted recordings were converted to single-channel 16-bit PCM WAV at 16~kHz, with Persian characters and spacing standardised by the \texttt{shekar} Normalizer~\citep{shekar2025joss}. Contributors were manually assigned by the authors to training, validation, and test partitions (142/26/30 of the 198 registered contributors), with gender balance and contributor diversity considered during assignment. Partitioning was performed at the contributor level, so the training, validation, and test splits are speaker-disjoint by construction, and these predefined splits are used throughout this study. Version 6 of the release carries an opaque contributor identifier on every clip, so the assignment is auditable rather than asserted: the released identifiers cover all 62{,}279 clips, and no contributor appears in more than one partition. They resolve to 190 contributors rather than 198 because the remaining eight registered contributors have no accepted recording in the release, so the released partitions hold 134/26/30 contributors.

\FloatBarrier
\section{Dataset Characteristics}
\label{sec:characterization}

\FloatBarrier
\subsection{Corpus Statistics and Comparison}
\label{sec:cv-comparison}
Partition-level statistics are reported in Table~\ref{tab:composition}. After \texttt{shekar} normalisation and tokenisation~\citep{shekar2025joss}, with punctuation-only tokens excluded, 626{,}370 tokens and 29{,}520 word types are counted. Evaluation-set out-of-vocabulary rates are estimated at 1.11--1.12\% against the training vocabulary.

Informal labels are assigned to 24.44\% of clips by the rule-based \texttt{shekar} InformalLanguageClassifier, with similar rates across partitions. With the library's ALBERT named-entity recogniser, 19{,}112 mentions are detected in 23.7\% of clips, predominantly locations and dates. These annotations are used descriptively and for evaluation strata, never as training labels. Neither register classification nor entity recognition has been validated against human labels for this corpus; the recording-acceptance agreement study does not validate these annotations. Version 1.6.3 of \texttt{shekar} is used in the experimental pipeline. Among the 198 registered contributors, female and male gender were reported by 93 and 81, respectively; gender was not reported by 24. The release identifies 190 of them, the remainder having no accepted recording.

\begin{table}[htbp]
\centering
\caption{Neyshekar statistics, overall and by partition. Contributor counts are the distinct contributor identifiers present in the release. Partitions are speaker-disjoint by construction, and the released per-clip identifiers confirm that no contributor appears in more than one partition.}
\label{tab:composition}
\small
\setlength{\tabcolsep}{4pt}
\renewcommand{\arraystretch}{1.15}
\begin{lrbox}{\papertablebox}
\begin{tabular}{lrrrr}
\toprule
Statistic & Full & Train & Val & Test \\
\midrule
Utterances & 62,279 & 58,244 & 1,886 & 2,149 \\
Duration (h) & 99.02 & 91.99 & 3.14 & 3.88 \\
Contributors & 190 & 134 & 26 & 30 \\
Mean dur. (s) & 5.72 & 5.69 & 6.00 & 6.49 \\
Words & 626,370 & 585,022 & 19,187 & 22,161 \\
Word types & 29,520 & 29,129 & 5,868 & 6,118 \\
Distinct prompts & 34,541 & 33,357 & 1,844 & 2,101 \\
Informal (\%) & 24.44 & 24.23 & 27.57 & 27.55 \\
\bottomrule
\end{tabular}
\end{lrbox}
\ifdim\wd\papertablebox>\linewidth
  \resizebox{\linewidth}{!}{\usebox{\papertablebox}}
\else
  \usebox{\papertablebox}
\fi
\end{table}

\begin{figure}[t]
\centering
\experimentfigure{0.8\columnwidth}{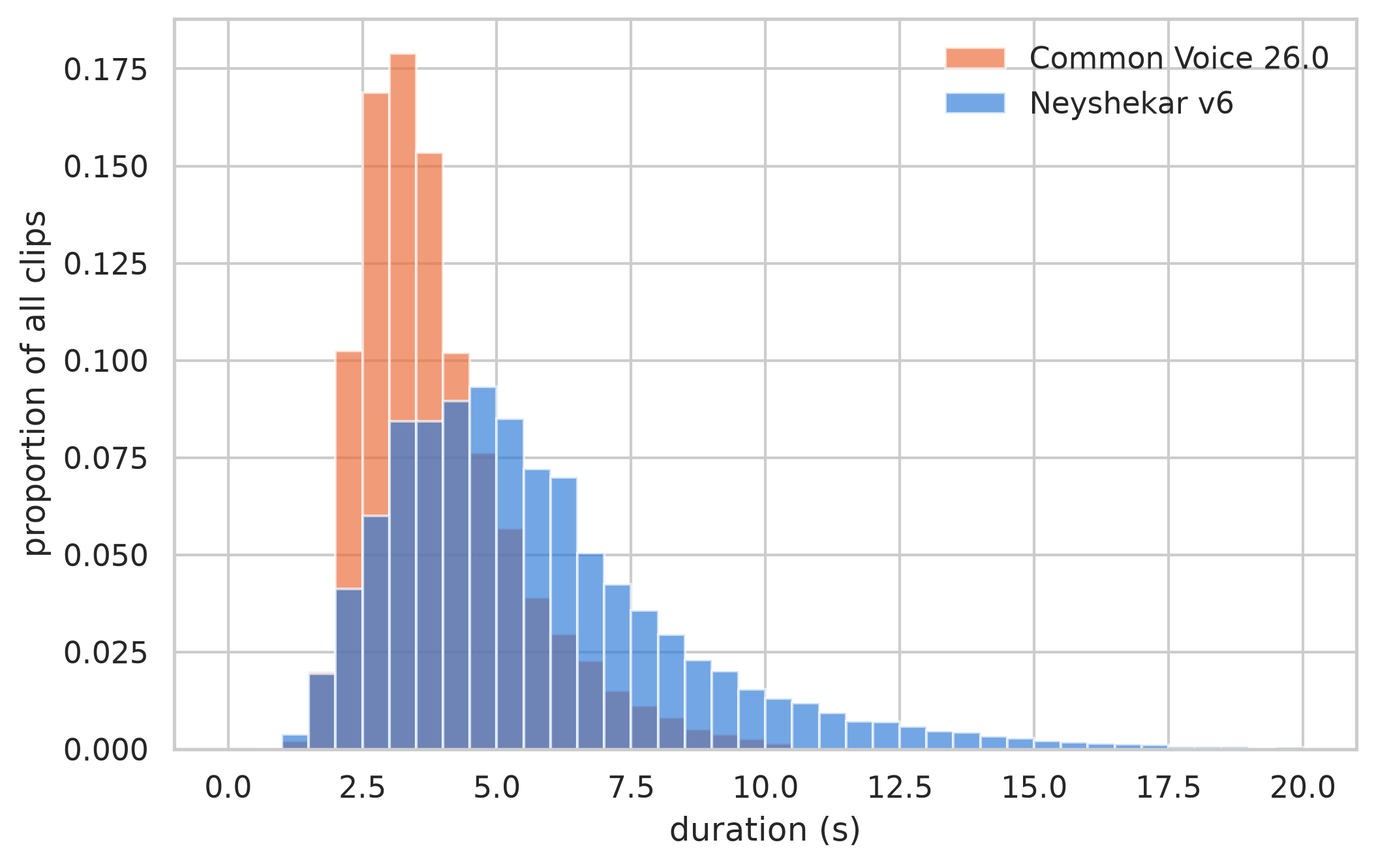}
\caption{Clip-duration distributions for Neyshekar v6 and validated Persian Common Voice 26.0 (proportion of all clips). The 88 Neyshekar clips and two Common Voice clips longer than 20 s (0.14\% and under 0.01\% of each corpus) are omitted rather than collected into the final bin.}
\label{fig:cv-duration}
\end{figure}

Longer utterances and greater informal and named-entity coverage are observed relative to validated Persian Common Voice 26.0~\citep{ardila2020commonvoice} under identical processing (Table~\ref{tab:cvcomparison}, Figure~\ref{fig:cv-duration}). Mean clip duration is 46\% longer for Neyshekar. At a matched 626{,}370 tokens, 18\% more word types are obtained than the average across five fixed-seed Common Voice samples. Common Voice NER is restricted to a fixed 62,279-clip sample. These distributional differences are used to motivate the mixture control; complementarity is not established by corpus statistics alone.

\begin{table}[htbp]
\centering
\caption{Neyshekar v6 vs.\ validated Persian Common Voice 26.0, identical processing.}
\label{tab:cvcomparison}
\small
\setlength{\tabcolsep}{4pt}
\renewcommand{\arraystretch}{1.15}
\begin{lrbox}{\papertablebox}
\begin{tabular}{lrr}
\toprule
Statistic & Neyshekar & Common Voice \\
\midrule
Clips & 62,279 & 341,657 \\
Duration (h) & 99.02 & 373.22 \\
Mean clip dur. (s) & 5.72 & 3.93 \\
Mean words/clip & 10.06 & 6.30 \\
Word types (matched tok.) & 29,520 & 24,973 \\
Informal clips (\%) & 24.44 & 16.08 \\
Entities / 100 words & 3.05 & 1.70 \\
Contributors / IDs & 190 & 4,338 \\
Readings per prompt & 1.80 & 6.58 \\
\bottomrule
\end{tabular}
\end{lrbox}
\ifdim\wd\papertablebox>\linewidth
  \resizebox{\linewidth}{!}{\usebox{\papertablebox}}
\else
  \usebox{\papertablebox}
\fi
\end{table}

\FloatBarrier
\subsection{Contributor Distribution}
\label{sec:contributors}
Recording load is spread unevenly, as in any crowdsourced collection. The 190 contributors supply a median of 129 clips (IQR 16--452, range 1--5,935) and a median of 12.2 minutes (range 0.04--457); the largest single contributor accounts for 7.7\% of audio and the ten largest for 35.6\%, with 23 contributors covering half the corpus (Table~\ref{tab:contributors}).

Concentration is nevertheless lower than in Persian Common Voice, whose 4,338 identifiers carry a Gini of 0.871 against 0.657 here, and 63.9\% of whose contributors supply under one minute in total against 17.9\%. Weighting by hours, the two corpora are closer than their nominal counts suggest: 70 effective contributors for Common Voice against 46 here, from 23$\times$ as many identifiers. Speaker variety remains a limitation of this corpus relative to Common Voice, and contributor-clustered intervals are reported for the Neyshekar test accordingly (Section~\ref{sec:statistics}).

\begin{table}[htbp]
\centering
\caption{Per-contributor recording load in Neyshekar v6 and validated Persian Common Voice 26.0. Effective contributors is the inverse Simpson index of the hours distribution: the number of equally contributing speakers that would produce the same concentration. Gini is computed over per-contributor hours, with 0 for an even split and 1 for total concentration. Common Voice is keyed on its released \texttt{client\_id}.}
\label{tab:contributors}
\small
\setlength{\tabcolsep}{4pt}
\renewcommand{\arraystretch}{1.15}
\begin{lrbox}{\papertablebox}
\begin{tabular}{@{}lrr@{}}
\toprule
Statistic & Neyshekar & Common Voice \\
\midrule
Contributors & 190 & 4,338 \\
Effective contributors & 45.6 & 70.1 \\
Clips, median [IQR] & 129 [16, 452] & 8 [5, 20] \\
Clips, range & 1--5,935 & 1--20,949 \\
Minutes, median [IQR] & 12.2 [1.6, 43.7] & 0.6 [0.4, 1.6] \\
Minutes, range & 0.04--457 & 0.03--1,359 \\
Gini (hours) & 0.657 & 0.871 \\
Top-10 share of hours (\%) & 35.6 & 31.7 \\
Contributors for 50\% of hours & 23 & 27 \\
Under 1 minute total (\%) & 17.9 & 63.9 \\
\bottomrule
\end{tabular}
\end{lrbox}
\ifdim\wd\papertablebox>\linewidth
  \resizebox{\linewidth}{!}{\usebox{\papertablebox}}
\else
  \usebox{\papertablebox}
\fi
\end{table}

\FloatBarrier
\subsection{Signal Quality}
\label{sec:signal-quality}
Recordings were made on unrestricted personal devices with no signal-to-noise threshold, so the delivered audio is characterised directly from the released waveforms (Table~\ref{tab:acoustic-quality}). All clips are 16~kHz mono. Median level is $-22.9$~dBFS and median silence ratio is 0.30, the latter reflecting the leading and trailing pauses of prompted reading rather than mid-utterance gaps.

Clipping is rare: 2,116 clips (3.4\%) contain a flattened run of at least three consecutive full-scale samples, and only 385 (0.62\%) have more than 0.1\% of their samples on the rail. Estimated SNR should be read with care: in 41,782 clips (67\%) the capture chain suppresses pauses to the quantisation floor, so a percentile-based estimate describes that gate rather than the room. Over the remaining 20,497 clips the median estimate is 49.8~dB (IQR 43.7--56.1), indicating that contributors largely followed the quiet-setting instruction. These measures are reference-free and are not a substitute for perceptual quality assessment.

\begin{table}[htbp]
\centering
\caption{Reference-free signal quality over all 62{,}279 released clips, measured on 25~ms frames with a 10~ms hop. Silence is a frame more than 30~dB below the clip's 95th-percentile frame level. SNR compares the 95th and 5th percentiles of frame power and is reported over the 20,497 clips whose pauses retain a measurable noise floor; the remaining clips have pauses suppressed to the quantisation floor by the capture chain, which inflates any such estimate. Clipping counts full-scale samples.}
\label{tab:acoustic-quality}
\small
\setlength{\tabcolsep}{4pt}
\renewcommand{\arraystretch}{1.15}
\begin{lrbox}{\papertablebox}
\begin{tabular}{@{}lrr@{}}
\toprule
Measure & Median [IQR] & Range \\
\midrule
Level (dBFS) & $-22.9$ [$-26.4$, $-18.5$] & $-56.8$ to $-6.7$ \\
Speech level (dBFS) & $-21.2$ [$-24.6$, $-16.7$] & $-53.8$ to $-5.8$ \\
Peak (dBFS) & $-3.1$ [$-7.5$, 0.0] & $-38.4$ to 0.0 \\
Estimated SNR (dB) & 49.8 [43.7, 56.1] & 16.3 to 74.1 \\
Silence ratio & 0.300 [0.231, 0.382] & 0.000 to 0.844 \\
Leading silence (s) & 0.64 [0.42, 0.85] & 0.00 to 5.29 \\
Trailing silence (s) & 0.40 [0.20, 0.62] & 0.00 to 14.79 \\
Clipped samples (\%) & 0.000 [0.000, 0.000] & 0.000 to 0.603 \\
\bottomrule
\end{tabular}
\end{lrbox}
\ifdim\wd\papertablebox>\linewidth
  \resizebox{\linewidth}{!}{\usebox{\papertablebox}}
\else
  \usebox{\papertablebox}
\fi
\end{table}

\FloatBarrier
\subsection{Prompt Overlap Across Partitions}
\label{sec:overlap}
In the 62{,}279 clips, 34{,}541 distinct prompts are represented (mean 1.80 readings/prompt, maximum 16). Prompt disjointness is not enforced in the predefined partitions: training transcripts are repeated in 63.8\% of test clips and 73.3\% of validation clips. A 777-clip text-disjoint test subset, with transcripts unseen in training, is therefore also released and evaluated throughout Section~\ref{sec:results}.

\FloatBarrier
\section{Evaluation Methodology}
\label{sec:experiments}

\FloatBarrier
\subsection{Evaluation Objectives}
The evaluation is designed to establish whether recognition accuracy is improved through Neyshekar training and whether the resulting gains are generalised beyond the source corpus. In-domain performance is first compared with duration-matched Common Voice training, after which transfer is assessed across architectures on text-disjoint prompts, Common Voice, and independent PSRB speech. Variation in recognition accuracy is then examined across register, named-entity presence, and utterance duration. The effects of corpus composition and training-data size are investigated through mixture and scaling experiments in which audio duration and optimisation budgets are controlled. Throughout the repeated-seed comparisons, subset selection is held fixed so that optimisation variability can be assessed independently of changes in training data.

\FloatBarrier
\subsection{Models and Training Configuration}
The effect of training-corpus selection is examined using two complementary ASR architectures: the encoder--decoder Whisper small model~\citep{radford2023whisper} and the CTC-based XLS-R-300M model~\citep{babu2022xlsr}. Adaptation gains are assessed relative to zero-shot Whisper small. Broader performance context is provided by zero-shot Whisper large-v3 and MMS-1B-all with its Persian adapter~\citep{pratap2024mms}, although corpus effects cannot be isolated through these comparisons because model size and pretraining differ.

Training-data availability is controlled through approximately 32-hour subsets of Neyshekar and Common Voice. Recordings exceeding 20 seconds are excluded before matching, and the resulting sample counts and durations are reported in Table~\ref{tab:training-durations}. The exclusion removes 77 of the 58,244 training clips (0.13\%, 0.47 hours), which reconciles the 91.99-hour training split of Table~\ref{tab:composition} with the 91.52-hour full manifest used for training. Subset membership and clip order are fixed with data seed 42, while optimisation variability is assessed separately with seeds 42, 43, and 44. Under the primary protocol, both architectures are trained for three epochs with an effective batch size of 64. Learning rates of $1\!\times\!10^{-5}$ and $3\!\times\!10^{-4}$ are used for Whisper and XLS-R, respectively, with 10\% linear warm-up and a frozen XLS-R convolutional encoder. The final checkpoint is evaluated in every condition, and a shared Neyshekar--Common Voice development population is used. External evaluation data are excluded from training and model selection.

Comparability across the fine-tuned CTC systems is maintained through a shared character inventory defined independently of corpus selection and test references. Persian and Arabic letters and marks, Latin letters, ASCII digits, and special symbols are included, with label coverage verified before training. Digit scripts are unified without number verbalisation in both training targets and evaluation text. The original MMS inventory is retained only for the zero-shot reference.

The contribution of additional data is distinguished from that of additional optimisation through complementary budget controls. The approximately 32-hour comparison is repeated at 1,425 updates, that is $3\lceil30{,}385/64\rceil$ for the larger of the two subsets, and corpus composition is examined by comparing 64-hour mixed and Neyshekar-only subsets at 2,382 updates. Scaling is evaluated using fixed nested subsets under both three-epoch and 1,425-update schedules. Three optimisation seeds are used for corpus and mixture comparisons; scaling is restricted to seed 42. These controls permit comparisons at equal update counts, without implying equal computational cost or audio exposure.

\begin{table}[htbp]
\centering
\caption{Actual post-filtering training durations from hash-verified frozen manifests. Nominal hours are used in condition labels.}
\label{tab:training-durations}
\small
\setlength{\tabcolsep}{4pt}
\renewcommand{\arraystretch}{1.15}
\begin{lrbox}{\papertablebox}
\begin{tabular}{@{}lrr@{}}
\toprule
Condition & Clips & Hours \\
\midrule
Neyshekar 32h & 20429 & 32.16 \\
CV 32h & 30385 & 32.16 \\
Mixed 32h & 25475 & 32.16 \\
Neyshekar 64h & 40876 & 64.32 \\
Mixed 64h & 50814 & 64.32 \\
Neyshekar 5h & 3224 & 5.00 \\
Neyshekar 10h & 6392 & 10.00 \\
Neyshekar 20h & 12746 & 20.00 \\
Neyshekar 40h & 25394 & 40.00 \\
Neyshekar full & 58167 & 91.52 \\
\bottomrule
\end{tabular}
\end{lrbox}
\ifdim\wd\papertablebox>\linewidth
  \resizebox{\linewidth}{!}{\usebox{\papertablebox}}
\else
  \usebox{\papertablebox}
\fi
\end{table}

Training and decoding were implemented with Transformers 5.16.1; pretrained checkpoint revisions are recorded in the run provenance. Each training run was executed on one GPU of a workstation equipped with two NVIDIA RTX A6000 GPUs, 256 GB of RAM, and 64 AMD Ryzen CPU cores.

\FloatBarrier
\subsection{Evaluation Protocol and Metrics}
A single full-test decoding is obtained for each system on Neyshekar and Common Voice using greedy decoding and a batch size of one. Persian transcription is explicitly selected for all Whisper systems (\texttt{language=fa}, \texttt{task=transcribe}); no additional hallucination filtering is applied. Valid audio frames are retained before CTC token collapse, and long-form Whisper decoding is applied to recordings exceeding 30 seconds. Evaluation audio is not truncated. No recordings exceed 30 seconds in either internal test set; the long-form path is required for eight of the 344 PSRB recordings. Predictions are saved with ordered reference identifiers, decoding provenance, and file hashes. All text-disjoint and stratified analyses are derived from these saved predictions. The text-disjoint subset is defined by the absence of each normalised test reference from the complete Neyshekar training split, and is labelled \emph{Disj.} in the result tables.

References and hypotheses are processed identically using \texttt{shekar} normalisation~\citep{shekar2025joss}, digit-script unification, removal of punctuation and non-spoken bidirectional formatting controls, and conversion of zero-width non-joiners to spaces. WER and CER are computed with jiwer 4.0.0 from corpus-level edit counts, including insertions. Normalised spaces are included in CER. Empty normalised references are excluded, and the number of scored recordings is reported. Register and entity labels are used as automatic descriptive annotations, without human validation; duration strata are defined as $<$4, 4--10, and $>$10 seconds.

External evaluation is performed on the 344-clip public PSRB sample (0.89 hours)~\citep{psrb2025}. A fixed revision is retained with audio and transcript hashes. Recordings are evaluated intact after correction of the documented CSV column-order mismatch. Findings are restricted to the public sample rather than extrapolated to the full benchmark, and pretraining overlap cannot be excluded.

\FloatBarrier
\subsection{Statistical Analysis}
\label{sec:statistics}
Training variability is summarised by the mean and sample standard deviation across the three optimisation seeds. Test-sampling uncertainty is estimated separately through 20{,}000 paired bootstrap resamples of recordings, with systems paired by saved test identifiers and references. Percentile 95\% confidence intervals are computed for WER and CER differences~\citep{bisani2004bootstrap}, and are interpreted conditional on the trained systems; training seeds are not resampled. Sensitivity to optimisation randomness is summarised by seed standard deviations, whereas sensitivity to the sampled test recordings is represented by bootstrap intervals. These quantities are reported separately because different sources of variability are represented; combined uncertainty intervals are not estimated. With only three seeds, the seed standard deviations are themselves imprecisely estimated, particularly for the more variable XLS-R conditions.

Neyshekar intervals are additionally estimated by resampling contributors rather than utterances, using the released per-clip identifiers (Section~\ref{sec:partitioning}). The Neyshekar test carries 30 contributors, so contributor-clustered intervals rest on few clusters and are correspondingly wider. Source recordings and speakers are not identified in the public PSRB labels, so dependence at these levels is not represented by utterance resampling. Condition-level comparisons are treated as exploratory, without adjustment for multiplicity.

\FloatBarrier
\section{Results and Discussion}
\label{sec:results}

\subsection{In-Domain Performance}
\label{sec:in-domain}
Lower in-domain error rates are obtained with Neyshekar training for both architectures. Under the three-epoch, approximately 32h comparison, mean Neyshekar WER is reduced by 9.46 percentage points for Whisper and 11.55 points for XLS-R, based on the displayed means (Table~\ref{tab:baselines}). Lower CER is also obtained, and zero is excluded from every paired WER and CER interval. The advantage does not rest on treating clips as independent: resampling the 30 test contributors instead of the 2,149 utterances leaves every per-seed WER interval clear of zero (Table~\ref{tab:clustered-intervals}). Training duration and preprocessing are held fixed, but prompt, speaker, and acoustic differences are not separately controlled.

Evidence of benefit beyond the Neyshekar partition is also obtained on the independent PSRB sample (Table~\ref{tab:external}). Mean WER is reduced from 51.61 to 43.42 for Whisper and from 64.47 to 56.54 for XLS-R. These gains are obtained on speech from an independent source, so they corroborate the in-domain result without relying on the contributor-level assignment of Section~\ref{sec:partitioning}: in-domain gains inflated by speaker overlap would not transfer to PSRB.

\begin{figure}[htbp]
\centering
\experimentfigure{0.78\columnwidth}{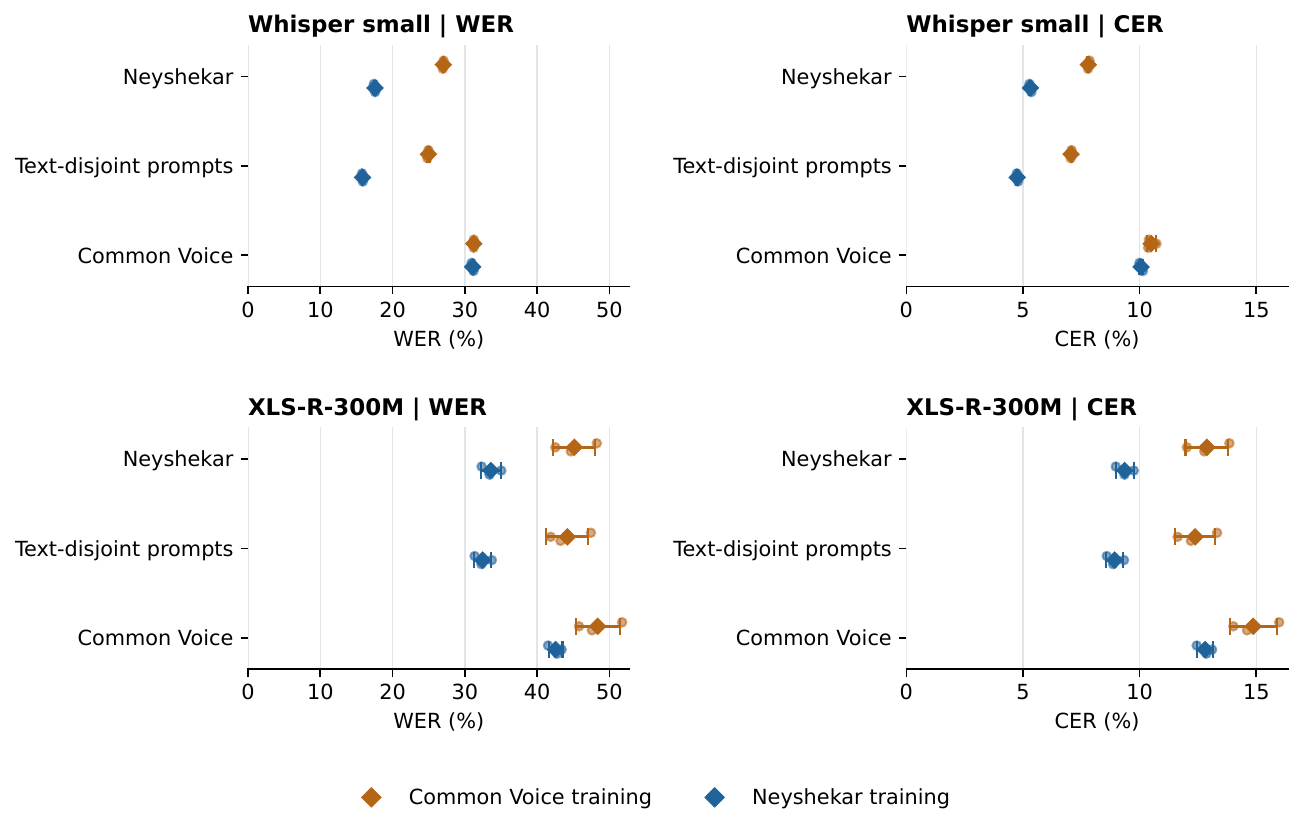}
\caption{Cross-corpus performance after three-epoch training on frozen approximately 32h subsets. WER and CER are shown separately. Small dots denote individual optimisation seeds; diamonds and whiskers denote the mean and sample SD across three seeds, not confidence intervals. Lower values indicate lower error rates.}
\label{fig:cross-corpus}
\end{figure}

Cross-corpus error patterns are visualised in Figure~\ref{fig:cross-corpus}.

\begin{table}[htbp]
\centering
\caption{Internal WER/CER (\%; lower is better). Fine-tuning is performed for three epochs on frozen approximately 32h subsets; values are reported as mean $\pm$ sample SD over seeds 42/43/44. N: full Neyshekar test (2,149 clips); Disj.: text-disjoint subset (777); CV: Common Voice test (10,752). Scale and pretraining are not controlled across zero-shot systems.}
\label{tab:baselines}
\small
\setlength{\tabcolsep}{4pt}
\renewcommand{\arraystretch}{1.15}
\begin{lrbox}{\papertablebox}
\begin{tabular}{@{}llrrrrrr@{}}
\toprule
System & Training & N WER & N CER & Disj. WER & Disj. CER & CV WER & CV CER \\
\midrule
Whisper small & Zero-shot & 62.67 & 23.51 & 57.73 & 20.90 & 90.75 & 47.30 \\
Whisper large-v3 & Zero-shot & 19.89 & 5.59 & 18.67 & 5.19 & 31.06 & 11.26 \\
MMS-1B-all & Zero-shot & 23.09 & 5.70 & 21.73 & 5.35 & 28.75 & 8.06 \\
Whisper small & CV 32h & $27.02 \pm 0.06$ & $7.80 \pm 0.05$ & $24.96 \pm 0.15$ & $7.06 \pm 0.02$ & $31.24 \pm 0.01$ & $10.49 \pm 0.20$ \\
Whisper small & Neyshekar 32h & $17.56 \pm 0.14$ & $5.32 \pm 0.04$ & $15.85 \pm 0.07$ & $4.75 \pm 0.03$ & $31.07 \pm 0.13$ & $10.06 \pm 0.07$ \\
XLS-R-300M & CV 32h & $45.14 \pm 2.90$ & $12.88 \pm 0.92$ & $44.18 \pm 2.90$ & $12.38 \pm 0.86$ & $48.38 \pm 3.06$ & $14.87 \pm 1.01$ \\
XLS-R-300M & Neyshekar 32h & $33.59 \pm 1.37$ & $9.36 \pm 0.39$ & $32.45 \pm 1.20$ & $8.93 \pm 0.37$ & $42.56 \pm 0.96$ & $12.81 \pm 0.33$ \\
\bottomrule
\end{tabular}
\end{lrbox}
\ifdim\wd\papertablebox>\linewidth
  \resizebox{\linewidth}{!}{\usebox{\papertablebox}}
\else
  \usebox{\papertablebox}
\fi
\end{table}

\subsection{Evaluation on Text-Disjoint Prompts}
\label{sec:overlap-results}
The in-domain advantage is retained after previously seen prompts are excluded. On the 777 text-disjoint clips, lower WER is obtained with Neyshekar training for both architectures, and the advantage is supported by all six per-seed WER intervals (Table~\ref{tab:disjoint-intervals}). Intrinsically lower difficulty of text-disjoint prompts cannot be inferred from the lower absolute WER on this subset. Register, length, and vocabulary are distributed differently from the remaining test clips, and a balanced test sample was not constructed.

\subsection{Cross-Corpus Transfer to Common Voice}
Similar Common Voice performance is obtained with Whisper after training on either duration-matched corpus: 31.07 WER with Neyshekar and 31.24 with Common Voice (Table~\ref{tab:baselines}). At the same time, reductions of 9.46 WER points on Neyshekar and 8.19 points on PSRB are obtained with Neyshekar training (Tables~\ref{tab:baselines} and~\ref{tab:external}). Thus, these gains are obtained without an observed increase in mean Common Voice WER under the three-epoch 32h protocol. The 0.17-point Common Voice difference is not resolved by the per-seed intervals, all of which include zero; statistical equivalence has not been tested.

For XLS-R, a Common Voice improvement is supported by every per-seed WER interval. Mean WER is reduced from 48.38 to 42.56, with improvements of 2.42--10.25 points across seeds. Individual seeds are reported in Table~\ref{tab:seed-variability}, and paired intervals in Table~\ref{tab:bootstrap-transfer}, both in Appendix~\ref{sec:appendix-tables}.

Intervals are conditional on the trained systems; training seeds are not resampled.

\subsection{External Evaluation on PSRB}
The PSRB gains previewed in Section~\ref{sec:in-domain} are reported in full in Table~\ref{tab:external}, alongside the zero-shot reference systems. A Neyshekar-training advantage is supported by every per-seed PSRB WER interval. Less conclusive evidence is obtained for Whisper CER, for which zero is included in two of three intervals.

\begin{figure}[htbp]
\centering
\experimentfigure{0.78\columnwidth}{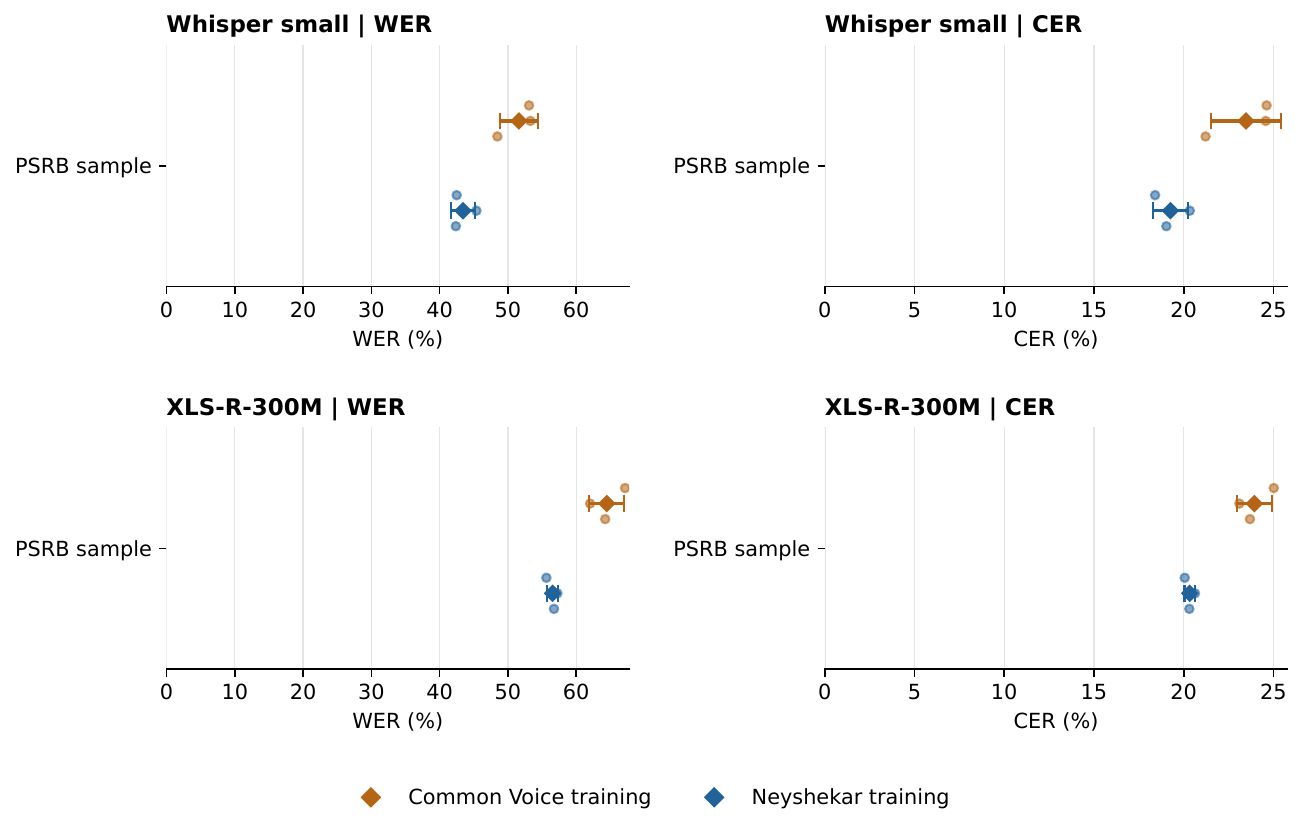}
\caption{External speech evaluation of the three-epoch, approximately 32h fine-tuned systems. Small dots denote individual seeds; diamonds and whiskers denote the three-seed mean and sample SD. PSRB is restricted to the 344-clip public sample. Lower WER/CER is better.}
\label{fig:external-results}
\end{figure}

External WER and CER are visualised in Figure~\ref{fig:external-results}.

\begin{table}[htbp]
\centering
\caption{Independent external WER/CER (\%). The 32h systems are trained for three epochs; the 64h systems are trained for 2,382 updates. Mean $\pm$ sample SD over three seeds. PSRB: 344 public-sample clips.}
\label{tab:external}
\small
\setlength{\tabcolsep}{4pt}
\renewcommand{\arraystretch}{1.15}
\begin{lrbox}{\papertablebox}
\begin{tabular}{@{}llrr@{}}
\toprule
System & Training & PSRB WER & PSRB CER \\
\midrule
Whisper small & Zero-shot & 93.20 & 56.56 \\
Whisper large-v3 & Zero-shot & 37.39 & 17.08 \\
MMS-1B-all & Zero-shot & 45.05 & 16.16 \\
Whisper small & CV 32h & $51.61 \pm 2.73$ & $23.47 \pm 1.95$ \\
Whisper small & Neyshekar 32h & $43.42 \pm 1.71$ & $19.26 \pm 0.99$ \\
Whisper small & Neyshekar 64h & $42.01 \pm 0.99$ & $19.44 \pm 1.43$ \\
Whisper small & Mixed 64h & $41.54 \pm 1.26$ & $18.63 \pm 1.37$ \\
XLS-R-300M & CV 32h & $64.47 \pm 2.58$ & $23.94 \pm 0.98$ \\
XLS-R-300M & Neyshekar 32h & $56.54 \pm 0.82$ & $20.33 \pm 0.29$ \\
XLS-R-300M & Neyshekar 64h & $45.39 \pm 1.28$ & $17.08 \pm 0.46$ \\
XLS-R-300M & Mixed 64h & $49.29 \pm 1.44$ & $17.94 \pm 0.43$ \\
\bottomrule
\end{tabular}
\end{lrbox}
\ifdim\wd\papertablebox>\linewidth
  \resizebox{\linewidth}{!}{\usebox{\papertablebox}}
\else
  \usebox{\papertablebox}
\fi
\end{table}

Broad spontaneous-speech robustness is not established by these sample-level gains. Lower PSRB WER is obtained with zero-shot large-v3 than with the fine-tuned small models, while the lowest zero-shot CER is obtained with MMS. These rankings should be interpreted under the declared decoding protocol and the limited coverage of the public sample.

\subsection{Sensitivity to the Update Budget}
The Neyshekar-training advantage is retained when optimiser updates are also matched. Under the 1,425-update comparison, lower Neyshekar and Common Voice WER/CER are obtained for both architectures (Table~\ref{tab:updates}). For Whisper, the Common Voice WER reduction is now supported by every seed's interval (Table~\ref{tab:update-intervals}), unlike in the equal-epoch comparison. Cross-budget comparisons are treated as sensitivity analyses.

\begin{table}[htbp]
\centering
\caption{Approximately 32h subsets under a common budget of 1,425 updates; WER/CER (\%), mean $\pm$ sample SD over three seeds. Audio exposure and FLOPs are not equalised by matching updates.}
\label{tab:updates}
\small
\setlength{\tabcolsep}{4pt}
\renewcommand{\arraystretch}{1.15}
\begin{lrbox}{\papertablebox}
\begin{tabular}{@{}llrrrrrr@{}}
\toprule
System & Training & N WER & N CER & Disj. WER & Disj. CER & CV WER & CV CER \\
\midrule
Whisper small & CV 32h & $27.01 \pm 0.03$ & $7.80 \pm 0.04$ & $24.92 \pm 0.11$ & $7.05 \pm 0.02$ & $31.18 \pm 0.09$ & $10.49 \pm 0.16$ \\
Whisper small & Neyshekar 32h & $16.43 \pm 0.03$ & $4.98 \pm 0.02$ & $14.79 \pm 0.13$ & $4.43 \pm 0.04$ & $30.24 \pm 0.08$ & $9.77 \pm 0.04$ \\
Whisper small & Mixed 32h & $18.45 \pm 0.08$ & $5.58 \pm 0.01$ & $16.44 \pm 0.07$ & $4.97 \pm 0.01$ & $29.23 \pm 0.04$ & $9.62 \pm 0.04$ \\
XLS-R-300M & CV 32h & $44.19 \pm 3.89$ & $12.54 \pm 1.31$ & $43.15 \pm 4.17$ & $12.10 \pm 1.37$ & $47.03 \pm 3.82$ & $14.42 \pm 1.33$ \\
XLS-R-300M & Neyshekar 32h & $26.87 \pm 2.06$ & $7.36 \pm 0.60$ & $25.96 \pm 2.01$ & $7.00 \pm 0.59$ & $37.88 \pm 1.85$ & $11.31 \pm 0.58$ \\
XLS-R-300M & Mixed 32h & $34.28 \pm 2.42$ & $9.43 \pm 0.72$ & $33.14 \pm 2.30$ & $8.96 \pm 0.67$ & $40.56 \pm 2.35$ & $12.14 \pm 0.77$ \\
\bottomrule
\end{tabular}
\end{lrbox}
\ifdim\wd\papertablebox>\linewidth
  \resizebox{\linewidth}{!}{\usebox{\papertablebox}}
\else
  \usebox{\papertablebox}
\fi
\end{table}

\subsection{Stratified Recognition Performance}
\label{sec:stratified-results}
Informal speech is not recognised less accurately by every system. After Neyshekar training, higher informal than formal WER is obtained with Whisper, whereas the reverse is observed with XLS-R (Table~\ref{tab:stratified}). Higher informal WER is observed for both Common Voice-trained systems. Across all four conditions, higher WER is obtained on entity-containing clips than on entity-free clips, and on short clips than on long clips (Table~\ref{tab:duration-strata}). These differences are descriptive because duration, register, and vocabulary are not independently controlled.

\begin{table}[htbp]
\centering
\caption{Neyshekar test strata: mean WER / CER (\%) across three seeds under three-epoch 32h training. Register and entity labels are assigned automatically without human validation; comparisons between strata are treated as descriptive.}
\label{tab:stratified}
\small
\setlength{\tabcolsep}{4pt}
\renewcommand{\arraystretch}{1.15}
\begin{lrbox}{\papertablebox}
\begin{tabular}{@{}llrrrr@{}}
\toprule
System & Training & Formal & Informal & No entity & Entity present \\
\midrule
Whisper small & CV 32h & 26.19 / 7.56 & 29.58 / 8.58 & 25.69 / 7.24 & 29.29 / 8.73 \\
Whisper small & Neyshekar 32h & 17.33 / 5.23 & 18.29 / 5.61 & 16.65 / 4.97 & 19.12 / 5.89 \\
XLS-R-300M & CV 32h & 43.97 / 12.48 & 48.77 / 14.18 & 44.27 / 12.51 & 46.63 / 13.49 \\
XLS-R-300M & Neyshekar 32h & 34.00 / 9.47 & 32.31 / 9.00 & 32.55 / 8.95 & 35.35 / 10.03 \\
\bottomrule
\end{tabular}
\end{lrbox}
\ifdim\wd\papertablebox>\linewidth
  \resizebox{\linewidth}{!}{\usebox{\papertablebox}}
\else
  \usebox{\papertablebox}
\fi
\end{table}

\subsection{Named-Entity Error Analysis}
Higher error rates within entity spans are also observed through reference-side substitution/deletion attribution. For three-epoch Neyshekar-trained Whisper, entity-word substitution/deletion rate is estimated at 22.18\%, compared with 15.16\% outside entities; the same ordering is observed for XLS-R (Table~\ref{tab:entity-attribution}). Insertions are excluded, so these values should not be interpreted as entity WER/CER. Thirteen mentions are left unmatched on the Neyshekar test, and the non-entity category cannot be guaranteed to be entity-free. Spaces are excluded from character attribution but included in overall CER.

\subsection{Effects of Training Data Composition}
\label{sec:extended}
Corpus composition is compared under matched audio duration and optimiser updates. The Mixed 32h condition is constructed from approximately 16.08 hours per corpus: 10,275 Neyshekar clips and 15,200 Common Voice clips. In the primary control, 64.32 hours of mixed audio (50,814 clips) are compared with 64.32 hours of Neyshekar-only audio (40,876 clips), with durations matched to within one second. The mixed condition is exactly the union of the two approximately 32-hour subsets used above (20{,}429 + 30{,}385 = 50{,}814 clips). Both conditions are trained for 2,382 updates with effective batch size 64 and seeds 42/43/44. The budget is fixed at $3\lceil50{,}814/64\rceil$; equal FLOPs or audio exposure are not implied by equal updates.

A cross-corpus tradeoff is observed when Whisper is trained on the mixture. Common Voice WER is reduced from 26.88 to 25.54, but Neyshekar WER is increased from 12.92 to 14.49 (Table~\ref{tab:mixture}). Both effects are supported by every seed's interval. A small PSRB WER reduction is observed, but the difference is not resolved by the per-seed intervals.

For XLS-R, lower mean WER is obtained with Neyshekar-only training on Neyshekar, Common Voice, and PSRB. On Neyshekar, WER is increased from 17.89 to 24.75 under mixing, but this comparison also involves a reduction from approximately 64 to 32 hours of in-domain audio. The Common Voice and PSRB results are therefore more informative about transfer beyond the training corpus. Neyshekar-only training is favoured by all per-seed WER intervals, although the Common Voice seed-43 interval is bounded very close to zero. A general complementarity claim is therefore not supported for this mixture and budget; other ratios and schedules cannot be excluded on this basis.

Because the mixture is the union of the two approximately 32-hour subsets, the three-epoch mixed condition is precisely the 32h Neyshekar subset held fixed with the Common Voice subset added to it. For Whisper, Neyshekar WER is reduced from 17.56 to 14.46 and Common Voice WER from 31.07 to 25.53 (Tables~\ref{tab:baselines} and~\ref{tab:mixture-epochs}). Improvement is therefore observed with the additional training resources, but it cannot be attributed specifically to corpus complementarity because both data volume and the number of updates are increased.

\begin{table}[htbp]
\centering
\caption{Duration- and update-matched control: approximately 64.32h and 2,382 updates per run. WER/CER (\%), mean $\pm$ sample SD over three seeds. Approximately 32h from each source are included in the mixture.}
\label{tab:mixture}
\small
\setlength{\tabcolsep}{4pt}
\renewcommand{\arraystretch}{1.15}
\begin{lrbox}{\papertablebox}
\begin{tabular}{@{}llrrrrrr@{}}
\toprule
System & Training & N WER & N CER & Disj. WER & Disj. CER & CV WER & CV CER \\
\midrule
Whisper small & Neyshekar 64h & $12.92 \pm 0.07$ & $3.91 \pm 0.04$ & $11.98 \pm 0.17$ & $3.62 \pm 0.05$ & $26.88 \pm 0.08$ & $8.61 \pm 0.01$ \\
Whisper small & Mixed 64h & $14.49 \pm 0.08$ & $4.35 \pm 0.01$ & $12.89 \pm 0.22$ & $3.92 \pm 0.03$ & $25.54 \pm 0.14$ & $8.31 \pm 0.05$ \\
XLS-R-300M & Neyshekar 64h & $17.89 \pm 1.20$ & $4.75 \pm 0.31$ & $16.84 \pm 1.07$ & $4.45 \pm 0.27$ & $30.33 \pm 0.97$ & $8.95 \pm 0.34$ \\
XLS-R-300M & Mixed 64h & $24.75 \pm 1.28$ & $6.55 \pm 0.37$ & $23.36 \pm 1.27$ & $6.07 \pm 0.36$ & $31.88 \pm 1.59$ & $9.37 \pm 0.55$ \\
\bottomrule
\end{tabular}
\end{lrbox}
\ifdim\wd\papertablebox>\linewidth
  \resizebox{\linewidth}{!}{\usebox{\papertablebox}}
\else
  \usebox{\papertablebox}
\fi
\end{table}

\begin{table}[htbp]
\centering
\caption{Three-epoch mixture sensitivity results (three seeds; mean $\pm$ sample SD). Durations and optimiser updates are not held constant across conditions; complementarity cannot be isolated from this table alone.}
\label{tab:mixture-epochs}
\small
\setlength{\tabcolsep}{4pt}
\renewcommand{\arraystretch}{1.15}
\begin{lrbox}{\papertablebox}
\begin{tabular}{@{}llrrrrrr@{}}
\toprule
System & Training & N WER & N CER & Disj. WER & Disj. CER & CV WER & CV CER \\
\midrule
Whisper small & Neyshekar 64h & $13.55 \pm 0.09$ & $4.11 \pm 0.01$ & $12.34 \pm 0.20$ & $3.74 \pm 0.06$ & $27.29 \pm 0.06$ & $8.84 \pm 0.05$ \\
Whisper small & Mixed 64h & $14.46 \pm 0.10$ & $4.34 \pm 0.01$ & $12.88 \pm 0.17$ & $3.92 \pm 0.04$ & $25.53 \pm 0.10$ & $8.37 \pm 0.07$ \\
Whisper small & Mixed 32h & $18.93 \pm 0.07$ & $5.80 \pm 0.08$ & $16.96 \pm 0.15$ & $5.12 \pm 0.01$ & $29.58 \pm 0.08$ & $9.75 \pm 0.11$ \\
XLS-R-300M & Neyshekar 64h & $20.20 \pm 0.30$ & $5.41 \pm 0.13$ & $19.07 \pm 0.25$ & $5.04 \pm 0.10$ & $32.06 \pm 0.34$ & $9.47 \pm 0.13$ \\
XLS-R-300M & Mixed 64h & $24.65 \pm 0.47$ & $6.54 \pm 0.13$ & $23.24 \pm 0.67$ & $6.05 \pm 0.20$ & $31.85 \pm 0.72$ & $9.36 \pm 0.24$ \\
XLS-R-300M & Mixed 32h & $36.46 \pm 3.46$ & $10.13 \pm 1.05$ & $35.66 \pm 3.54$ & $9.77 \pm 1.02$ & $42.34 \pm 3.30$ & $12.72 \pm 1.12$ \\
\bottomrule
\end{tabular}
\end{lrbox}
\ifdim\wd\papertablebox>\linewidth
  \resizebox{\linewidth}{!}{\usebox{\papertablebox}}
\else
  \usebox{\papertablebox}
\fi
\end{table}

The corresponding paired intervals are reported in Table~\ref{tab:bootstrap-mixture} and plotted per seed in Figure~\ref{fig:mixture-results}. Because the comparisons are exploratory and are not adjusted for multiplicity, the borderline XLS-R seed-43 Common Voice interval should not be interpreted as strong family-wise evidence.

\subsection{Effects of Training Data Size}
\begin{figure}[htbp]
\centering
\experimentfigure{0.92\columnwidth}{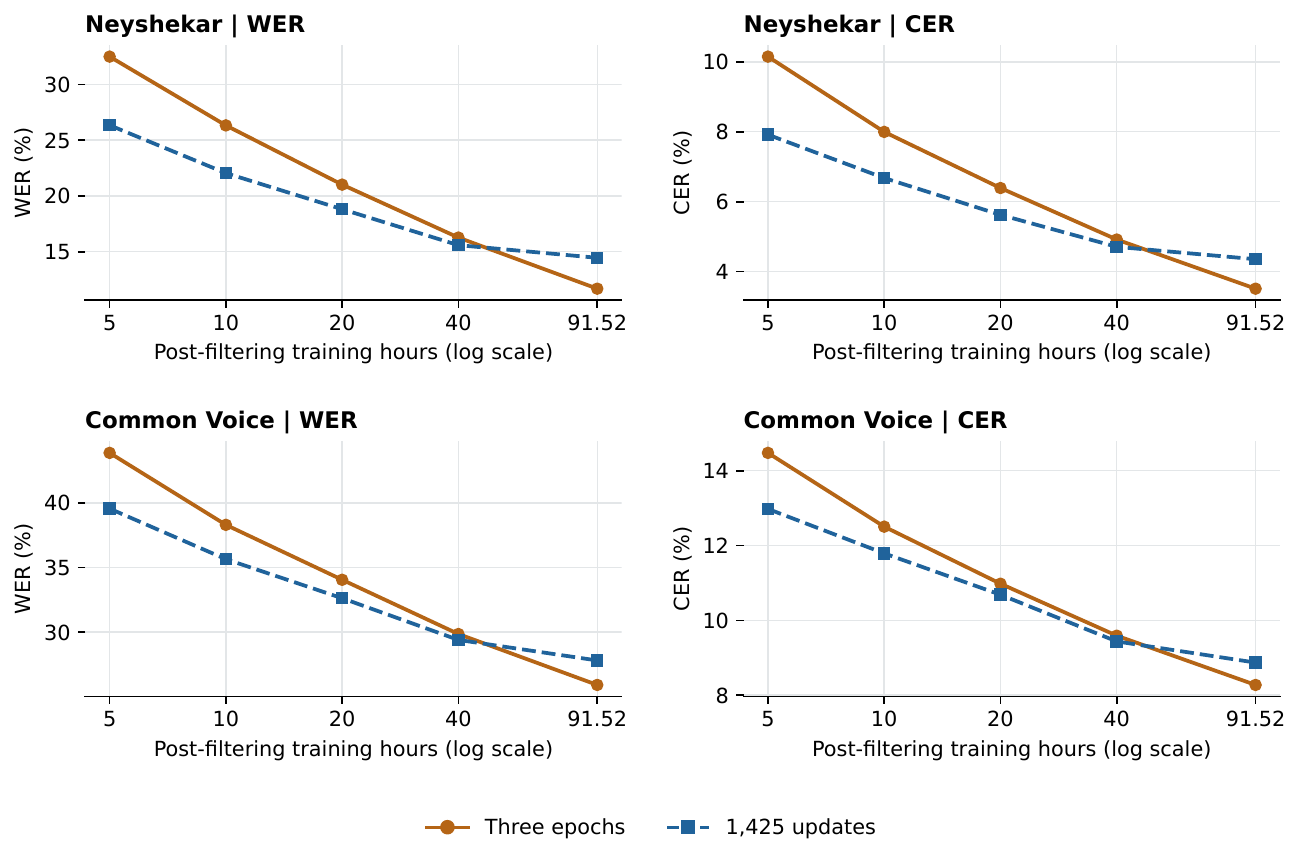}
\caption{Whisper small scaling under three epochs and a fixed 1,425-update budget. Actual post-filtering durations from frozen manifests are plotted on a logarithmic horizontal axis. Only optimisation seed 42 is represented; uncertainty bands are therefore not provided. Equal updates do not imply equal FLOPs or audio exposure.}
\label{fig:scaling-results}
\end{figure}

Improvements from additional distinct training data are retained at fixed updates. From 5h to the full 91.52h eligible split, Neyshekar WER is reduced from 26.34 to 14.46 at 1,425 updates; lower Common Voice error rates are also obtained (Table~\ref{tab:scaling}). Still lower full-split errors are obtained under the longer three-epoch schedule. The crossover in Figure~\ref{fig:scaling-results} follows from what a fixed budget buys at each size: 1,425 updates is approximately 28 epochs over the 5h subset but only 1.6 over the full 91.52h split, so the fixed budget exceeds three epochs at every size up to 40h and falls short of it at the full split. A benefit from larger training subsets is thus observed without increasing the number of optimiser updates. This control does not equate audio exposure or computational cost, and subset composition is varied along with size. Saturation is not established by these single-seed curves. Three-epoch mixture and 32h controls are reported in Tables~\ref{tab:mixture-epochs} and~\ref{tab:updates}, respectively.

\begin{table}[htbp]
\centering
\caption{Whisper small scaling, optimisation seed 42 only. Nominal hours (full training split after filtering: 91.52h); WER/CER (\%). Frozen nested subsets are shared across the two budgets.}
\label{tab:scaling}
\small
\setlength{\tabcolsep}{4pt}
\renewcommand{\arraystretch}{1.15}
\begin{lrbox}{\papertablebox}
\begin{tabular}{@{}rlrrrrrr@{}}
\toprule
Hours & Budget & N WER & N CER & Disj. WER & Disj. CER & CV WER & CV CER \\
\midrule
5 & 3 epochs & 32.49 & 10.15 & 29.48 & 8.98 & 43.87 & 14.48 \\
5 & 1,425 updates & 26.34 & 7.92 & 24.30 & 7.21 & 39.56 & 12.98 \\
10 & 3 epochs & 26.32 & 8.00 & 23.89 & 7.19 & 38.30 & 12.50 \\
10 & 1,425 updates & 22.05 & 6.68 & 19.80 & 5.92 & 35.65 & 11.79 \\
20 & 3 epochs & 21.02 & 6.39 & 18.53 & 5.61 & 34.05 & 10.98 \\
20 & 1,425 updates & 18.81 & 5.62 & 16.70 & 4.86 & 32.63 & 10.69 \\
40 & 3 epochs & 16.27 & 4.92 & 14.55 & 4.41 & 29.86 & 9.59 \\
40 & 1,425 updates & 15.59 & 4.71 & 14.14 & 4.22 & 29.40 & 9.44 \\
91.52 & 3 epochs & 11.69 & 3.51 & 10.66 & 3.18 & 25.92 & 8.27 \\
91.52 & 1,425 updates & 14.46 & 4.35 & 12.67 & 3.83 & 27.82 & 8.87 \\
\bottomrule
\end{tabular}
\end{lrbox}
\ifdim\wd\papertablebox>\linewidth
  \resizebox{\linewidth}{!}{\usebox{\papertablebox}}
\else
  \usebox{\papertablebox}
\fi
\end{table}

\FloatBarrier
\needspace{6\baselineskip}
\subsection{Implications for Persian ASR}
\label{sec:implications}
Neyshekar is supported as a source of adaptation data for the evaluated architectures, particularly when recognition of its read-speech distribution is required. Benefits from additional distinct training data are also observed under a fixed update budget. Corpus mixing should nevertheless be assessed against the intended evaluation domain: a Common Voice improvement is obtained for Whisper at the cost of in-domain accuracy, whereas a consistent mixing benefit is not established for XLS-R. These controlled comparisons should not be interpreted as deployment rankings. Lower PSRB WER is obtained with zero-shot Whisper large-v3 than with the fine-tuned small models, and adaptation of larger models has not been evaluated here.

\FloatBarrier
\section{Conclusion}
\label{sec:conclusion}
Neyshekar is released with 99.02 hours of Persian read speech, curated register and named-entity coverage, item-level rater labels, and a text-disjoint test subset. What the controlled comparisons support, and where they stop short, is set out in Section~\ref{sec:implications}. The corpus, the hash-verified training manifests, and the saved decodings are released together so that work beyond aggregate in-domain accuracy is enabled: per-condition error analysis, alternative mixture ratios and schedules, and evaluation on speech beyond prompted reading.

\FloatBarrier
\section*{Limitations}
\label{sec:limitations}
The principal limitations concern contributor breadth and the predominance of generated prompts. The corpus draws on 190 contributors against 4{,}338 identifiers in Persian Common Voice, and its test partition holds 30, so contributor-clustered intervals are estimated from few clusters and generalisation to unseen speakers remains uncertain. The corpus is restricted to prompted read speech; naturalness and register representativeness have not been established through a dedicated human evaluation, and register/entity annotations remain automatically derived. Possible overlap among the web sources used to pretrain the prompt generators and ASR models cannot be excluded. Text-disjoint evaluation is restricted to a smaller sample, without balancing against seen prompts. Training variability is assessed with three optimisation seeds, and scaling with only one. PSRB evaluation is restricted to the public sample without source-recording grouping. Broad conversational or far-field robustness is not established, and pretraining contamination cannot be excluded. Decoding was not tuned on the external test. Corpus effects cannot be attributed specifically to entity curation or informal prompts without corresponding ablations.

\FloatBarrier
\section*{Ethics Statement}
All recordings were voluntarily contributed by adults for inclusion in the publicly released speech dataset. Contributors were informed of the purpose and public release of their recordings, and consent was obtained before submission. Personally identifying information was not included in the released metadata.

\FloatBarrier
\section*{Generative AI Use}
Generative AI tools were used during manuscript preparation to assist with language editing and improving the clarity of portions of the text. All AI-assisted content was reviewed and revised by the authors. The study design, data collection, analyses, interpretation of results, and conclusions remained under the authors' responsibility, and the authors take full responsibility for the accuracy and integrity of the final manuscript.

\FloatBarrier
\section*{Data and Code Availability}
The Neyshekar corpus is released under the CC0 1.0 Universal licence and archived on Zenodo~\citep{neyshekar_data} under the concept DOI \url{https://doi.org/10.5281/zenodo.18073632}. Version 6 is evaluated in this study. The project repository and code are available at \url{https://github.com/amirivojdan/neyshekar}.

\FloatBarrier
\section*{Acknowledgements}
Neyshekar was built one recording at a time by the volunteers who lent it their voices, and they are gratefully acknowledged. Thanks are also owed to the annotators whose careful listening made the reliability study possible.

\FloatBarrier
\bibliography{custom}

\appendix
\setcounter{table}{0}
\setcounter{figure}{0}
\renewcommand{\thetable}{A\arabic{table}}
\renewcommand{\thefigure}{A\arabic{figure}}

\section{Supplementary Results}
\label{sec:appendix-tables}
Per-seed results and the paired bootstrap intervals cited in Section~\ref{sec:results} are collected here. Every interval is conditional on the trained systems and is not adjusted for multiplicity. Intervals are computed by resampling utterances within each test, except in Table~\ref{tab:clustered-intervals}, which contrasts utterance resampling with resampling whole contributors on the Neyshekar test.

\begin{table}[htbp]
\centering
\caption{Individual optimisation seeds for the three-epoch matched-duration comparison. WER/CER (\%).}
\label{tab:seed-variability}
\small
\setlength{\tabcolsep}{4pt}
\renewcommand{\arraystretch}{1.15}
\begin{lrbox}{\papertablebox}
\begin{tabular}{@{}llrrrrrrr@{}}
\toprule
System & Training & Seed & N WER & N CER & Disj. WER & Disj. CER & CV WER & CV CER \\
\midrule
Whisper small & CV 32h & 42 & 27.05 & 7.85 & 24.96 & 7.08 & 31.23 & 10.40 \\
Whisper small & CV 32h & 43 & 27.05 & 7.76 & 25.11 & 7.05 & 31.25 & 10.72 \\
Whisper small & CV 32h & 44 & 26.96 & 7.79 & 24.82 & 7.05 & 31.22 & 10.37 \\
Whisper small & Neyshekar 32h & 42 & 17.41 & 5.28 & 15.77 & 4.73 & 30.93 & 9.99 \\
Whisper small & Neyshekar 32h & 43 & 17.69 & 5.32 & 15.88 & 4.73 & 31.08 & 10.05 \\
Whisper small & Neyshekar 32h & 44 & 17.59 & 5.36 & 15.88 & 4.79 & 31.20 & 10.14 \\
XLS-R-300M & CV 32h & 42 & 48.25 & 13.85 & 47.44 & 13.32 & 51.76 & 15.98 \\
XLS-R-300M & CV 32h & 43 & 42.52 & 12.03 & 41.88 & 11.63 & 45.80 & 14.02 \\
XLS-R-300M & CV 32h & 44 & 44.66 & 12.77 & 43.23 & 12.20 & 47.58 & 14.60 \\
XLS-R-300M & Neyshekar 32h & 42 & 32.31 & 8.98 & 31.35 & 8.59 & 41.51 & 12.45 \\
XLS-R-300M & Neyshekar 32h & 43 & 35.04 & 9.75 & 33.73 & 9.33 & 43.38 & 13.10 \\
XLS-R-300M & Neyshekar 32h & 44 & 33.41 & 9.35 & 32.29 & 8.86 & 42.78 & 12.87 \\
\bottomrule
\end{tabular}
\end{lrbox}
\ifdim\wd\papertablebox>\linewidth
  \resizebox{\linewidth}{!}{\usebox{\papertablebox}}
\else
  \usebox{\papertablebox}
\fi
\end{table}

\begin{table}[htbp]
\centering
\caption{Paired differences in percentage points (CV-trained minus Neyshekar-trained); percentile 95\% intervals from 20,000 resamples. Utterances are resampled within each test. Intervals are estimated conditional on each trained pair, without multiplicity adjustment.}
\label{tab:bootstrap-transfer}
\small
\setlength{\tabcolsep}{4pt}
\renewcommand{\arraystretch}{1.15}
\begin{lrbox}{\papertablebox}
\begin{tabular}{@{}llrrr@{}}
\toprule
System & Test & Seed & WER difference [95\% CI] & CER difference [95\% CI] \\
\midrule
Whisper small & Neyshekar & 42 & 9.641 [8.996, 10.291] & 2.577 [2.344, 2.820] \\
Whisper small & Neyshekar & 43 & 9.366 [8.712, 10.013] & 2.437 [2.214, 2.663] \\
Whisper small & Neyshekar & 44 & 9.370 [8.710, 10.031] & 2.428 [2.199, 2.662] \\
Whisper small & CV & 42 & 0.296 [-0.227, 0.903] & 0.404 [-0.064, 1.050] \\
Whisper small & CV & 43 & 0.169 [-0.310, 0.708] & 0.666 [0.165, 1.292] \\
Whisper small & CV & 44 & 0.020 [-0.469, 0.527] & 0.228 [-0.227, 0.788] \\
Whisper small & PSRB & 42 & 10.595 [3.806, 19.150] & 6.215 [0.863, 13.001] \\
Whisper small & PSRB & 43 & 7.898 [2.618, 14.666] & 4.233 [-0.710, 10.120] \\
Whisper small & PSRB & 44 & 6.091 [0.590, 12.879] & 2.184 [-2.998, 8.260] \\
XLS-R-300M & Neyshekar & 42 & 15.940 [15.232, 16.647] & 4.869 [4.664, 5.080] \\
XLS-R-300M & Neyshekar & 43 & 7.488 [6.767, 8.208] & 2.278 [2.074, 2.481] \\
XLS-R-300M & Neyshekar & 44 & 11.247 [10.561, 11.922] & 3.420 [3.218, 3.622] \\
XLS-R-300M & CV & 42 & 10.246 [9.817, 10.669] & 3.534 [3.401, 3.666] \\
XLS-R-300M & CV & 43 & 2.416 [1.995, 2.835] & 0.915 [0.781, 1.050] \\
XLS-R-300M & CV & 44 & 4.804 [4.395, 5.223] & 1.736 [1.609, 1.865] \\
XLS-R-300M & PSRB & 42 & 11.524 [10.230, 12.863] & 4.961 [4.462, 5.457] \\
XLS-R-300M & PSRB & 43 & 4.788 [3.465, 6.069] & 2.475 [1.992, 2.951] \\
XLS-R-300M & PSRB & 44 & 7.498 [6.295, 8.723] & 3.377 [2.892, 3.858] \\
\bottomrule
\end{tabular}
\end{lrbox}
\ifdim\wd\papertablebox>\linewidth
  \resizebox{\linewidth}{!}{\usebox{\papertablebox}}
\else
  \usebox{\papertablebox}
\fi
\end{table}

\begin{table}[htbp]
\centering
\caption{Mean reference-side substitution/deletion rates (\%) on the Neyshekar test, three-epoch 32h systems. Insertions are excluded; these rates should not be interpreted as entity WER/CER. Tokens are counted after scoring normalisation, including conversion of zero-width non-joiners to spaces, and therefore differ from the corpus-statistics tokenisation in Table~\ref{tab:composition}. Word spans: 1,426 entity and 23,292 non-entity tokens; 13 unmatched mentions are not included in labelled spans.}
\label{tab:entity-attribution}
\small
\setlength{\tabcolsep}{4pt}
\renewcommand{\arraystretch}{1.15}
\begin{lrbox}{\papertablebox}
\begin{tabular}{@{}llrrrr@{}}
\toprule
System & Training & Entity word & Other word & Entity char. & Other char. \\
\midrule
Whisper small & CV 32h & 33.24 & 24.10 & 7.38 & 5.62 \\
Whisper small & Neyshekar 32h & 22.18 & 15.16 & 5.66 & 4.14 \\
XLS-R-300M & CV 32h & 52.55 & 41.78 & 11.89 & 10.29 \\
XLS-R-300M & Neyshekar 32h & 41.28 & 30.54 & 9.71 & 7.74 \\
\bottomrule
\end{tabular}
\end{lrbox}
\ifdim\wd\papertablebox>\linewidth
  \resizebox{\linewidth}{!}{\usebox{\papertablebox}}
\else
  \usebox{\papertablebox}
\fi
\end{table}

\begin{table}[htbp]
\centering
\caption{Paired differences in percentage points (mixed minus Neyshekar-only); percentile 95\% intervals from 20,000 resamples. Utterances are resampled within each test. Intervals are estimated conditional on each trained pair, without multiplicity adjustment.}
\label{tab:bootstrap-mixture}
\small
\setlength{\tabcolsep}{4pt}
\renewcommand{\arraystretch}{1.15}
\begin{lrbox}{\papertablebox}
\begin{tabular}{@{}llrrr@{}}
\toprule
System & Test & Seed & WER difference [95\% CI] & CER difference [95\% CI] \\
\midrule
Whisper small & Neyshekar & 42 & 1.663 [1.237, 2.090] & 0.433 [0.293, 0.575] \\
Whisper small & Neyshekar & 43 & 1.388 [0.975, 1.807] & 0.399 [0.257, 0.540] \\
Whisper small & Neyshekar & 44 & 1.634 [1.225, 2.044] & 0.475 [0.336, 0.614] \\
Whisper small & CV & 42 & -1.401 [-1.716, -1.085] & -0.334 [-0.517, -0.110] \\
Whisper small & CV & 43 & -1.217 [-1.602, -0.786] & -0.317 [-0.574, 0.095] \\
Whisper small & CV & 44 & -1.417 [-1.807, -0.963] & -0.246 [-0.543, 0.182] \\
Whisper small & PSRB & 42 & -1.110 [-7.249, 3.477] & -2.356 [-7.988, 1.803] \\
Whisper small & PSRB & 43 & 0.813 [-2.415, 3.972] & 1.497 [-1.552, 5.079] \\
Whisper small & PSRB & 44 & -1.110 [-5.409, 2.347] & -1.550 [-6.391, 2.479] \\
XLS-R-300M & Neyshekar & 42 & 7.614 [7.053, 8.186] & 1.975 [1.805, 2.147] \\
XLS-R-300M & Neyshekar & 43 & 5.979 [5.425, 6.533] & 1.569 [1.411, 1.734] \\
XLS-R-300M & Neyshekar & 44 & 6.991 [6.414, 7.581] & 1.875 [1.700, 2.051] \\
XLS-R-300M & CV & 42 & 2.120 [1.770, 2.473] & 0.617 [0.508, 0.724] \\
XLS-R-300M & CV & 43 & 0.342 [0.003, 0.685] & -0.002 [-0.110, 0.106] \\
XLS-R-300M & CV & 44 & 2.185 [1.839, 2.537] & 0.648 [0.537, 0.760] \\
XLS-R-300M & PSRB & 42 & 4.633 [3.538, 5.733] & 1.023 [0.492, 1.537] \\
XLS-R-300M & PSRB & 43 & 2.994 [1.985, 4.019] & 0.755 [0.260, 1.216] \\
XLS-R-300M & PSRB & 44 & 4.065 [2.841, 5.289] & 0.812 [0.234, 1.376] \\
\bottomrule
\end{tabular}
\end{lrbox}
\ifdim\wd\papertablebox>\linewidth
  \resizebox{\linewidth}{!}{\usebox{\papertablebox}}
\else
  \usebox{\papertablebox}
\fi
\end{table}

\begin{table}[htbp]
\centering
\caption{Text-disjoint Neyshekar test, three-epoch 32h comparison. Differences are Common Voice-trained minus Neyshekar-trained error rates in percentage points; positive values favour Neyshekar training. Brackets give paired percentile 95\% intervals from 20,000 utterance resamples, conditional on each trained system pair and without multiplicity adjustment.}
\label{tab:disjoint-intervals}
\small
\setlength{\tabcolsep}{4pt}
\renewcommand{\arraystretch}{1.15}
\begin{lrbox}{\papertablebox}
\begin{tabular}{@{}lrrr@{}}
\toprule
System & Seed & $\Delta$ WER [95\% CI] & $\Delta$ CER [95\% CI] \\
\midrule
Whisper small & 42 & 9.188 [8.196, 10.190] & 2.345 [2.020, 2.679] \\
Whisper small & 43 & 9.229 [8.208, 10.253] & 2.325 [2.000, 2.661] \\
Whisper small & 44 & 8.932 [7.906, 9.945] & 2.259 [1.932, 2.591] \\
XLS-R-300M & 42 & 16.089 [14.934, 17.249] & 4.727 [4.399, 5.059] \\
XLS-R-300M & 43 & 8.152 [7.045, 9.279] & 2.299 [1.984, 2.613] \\
XLS-R-300M & 44 & 10.941 [9.825, 12.084] & 3.338 [3.016, 3.660] \\
\bottomrule
\end{tabular}
\end{lrbox}
\ifdim\wd\papertablebox>\linewidth
  \resizebox{\linewidth}{!}{\usebox{\papertablebox}}
\else
  \usebox{\papertablebox}
\fi
\end{table}

\begin{table}[htbp]
\centering
\caption{Common Voice test, 1,425-update 32h comparison. Differences are Common Voice-trained minus Neyshekar-trained error rates in percentage points; positive values favour Neyshekar training. Brackets give paired percentile 95\% intervals from 20,000 utterance resamples, conditional on each trained system pair and without multiplicity adjustment.}
\label{tab:update-intervals}
\small
\setlength{\tabcolsep}{4pt}
\renewcommand{\arraystretch}{1.15}
\begin{lrbox}{\papertablebox}
\begin{tabular}{@{}lrrr@{}}
\toprule
System & Seed & $\Delta$ WER [95\% CI] & $\Delta$ CER [95\% CI] \\
\midrule
Whisper small & 42 & 1.081 [0.553, 1.688] & 0.745 [0.274, 1.379] \\
Whisper small & 43 & 0.771 [0.343, 1.227] & 0.872 [0.449, 1.423] \\
Whisper small & 44 & 0.969 [0.473, 1.482] & 0.538 [0.137, 1.083] \\
XLS-R-300M & 42 & 7.551 [7.125, 7.979] & 2.443 [2.311, 2.577] \\
XLS-R-300M & 43 & 5.809 [5.374, 6.232] & 2.029 [1.895, 2.161] \\
XLS-R-300M & 44 & 14.095 [13.647, 14.542] & 4.839 [4.695, 4.985] \\
\bottomrule
\end{tabular}
\end{lrbox}
\ifdim\wd\papertablebox>\linewidth
  \resizebox{\linewidth}{!}{\usebox{\papertablebox}}
\else
  \usebox{\papertablebox}
\fi
\end{table}

\begin{table}[htbp]
\centering
\caption{Neyshekar test WER/CER (\%) by recording duration. Means are reported over seeds 42/43/44 for three-epoch 32h systems; all strata are selected from the same saved full-test predictions.}
\label{tab:duration-strata}
\small
\setlength{\tabcolsep}{4pt}
\renewcommand{\arraystretch}{1.15}
\begin{lrbox}{\papertablebox}
\begin{tabular}{@{}llrrr@{}}
\toprule
System & Training & $<4$s & 4--10s & $>10$s \\
\midrule
Whisper small & CV 32h & 28.20 / 8.66 & 28.02 / 8.19 & 23.54 / 6.38 \\
Whisper small & Neyshekar 32h & 19.30 / 6.21 & 18.68 / 5.70 & 13.55 / 3.92 \\
XLS-R-300M & CV 32h & 45.77 / 13.46 & 45.41 / 13.02 & 44.13 / 12.28 \\
XLS-R-300M & Neyshekar 32h & 34.48 / 10.07 & 33.65 / 9.42 & 33.09 / 8.95 \\
\bottomrule
\end{tabular}
\end{lrbox}
\ifdim\wd\papertablebox>\linewidth
  \resizebox{\linewidth}{!}{\usebox{\papertablebox}}
\else
  \usebox{\papertablebox}
\fi
\end{table}

\begin{table}[htbp]
\centering
\caption{Neyshekar test WER differences under two resampling units, for the three-epoch 32h comparison. Values are Common Voice-trained minus Neyshekar-trained WER in percentage points; positive favours Neyshekar training. Utterance resampling treats each clip as independent, whereas contributor resampling draws whole contributors and so reflects uncertainty about unseen speakers. Both use 20,000 paired resamples and percentile 95\% intervals, conditional on each trained pair and without multiplicity adjustment. The test partition holds 30 contributors.}
\label{tab:clustered-intervals}
\small
\setlength{\tabcolsep}{4pt}
\renewcommand{\arraystretch}{1.15}
\begin{lrbox}{\papertablebox}
\begin{tabular}{@{}llrr@{}}
\toprule
System & Seed & $\Delta$ WER, utterances & $\Delta$ WER, contributors \\
\midrule
Whisper small & 42 & 9.64 [9.00, 10.29] & 9.64 [9.03, 10.37] \\
Whisper small & 43 & 9.37 [8.71, 10.01] & 9.37 [8.59, 9.98] \\
Whisper small & 44 & 9.37 [8.71, 10.03] & 9.37 [8.69, 10.06] \\
XLS-R-300M & 42 & 15.94 [15.23, 16.65] & 15.94 [14.74, 16.90] \\
XLS-R-300M & 43 & 7.49 [6.77, 8.21] & 7.49 [6.21, 8.63] \\
XLS-R-300M & 44 & 11.25 [10.56, 11.92] & 11.25 [10.21, 12.16] \\
\bottomrule
\end{tabular}
\end{lrbox}
\ifdim\wd\papertablebox>\linewidth
  \resizebox{\linewidth}{!}{\usebox{\papertablebox}}
\else
  \usebox{\papertablebox}
\fi
\end{table}

\begin{figure}[htbp]
\centering
\experimentfigure{\columnwidth}{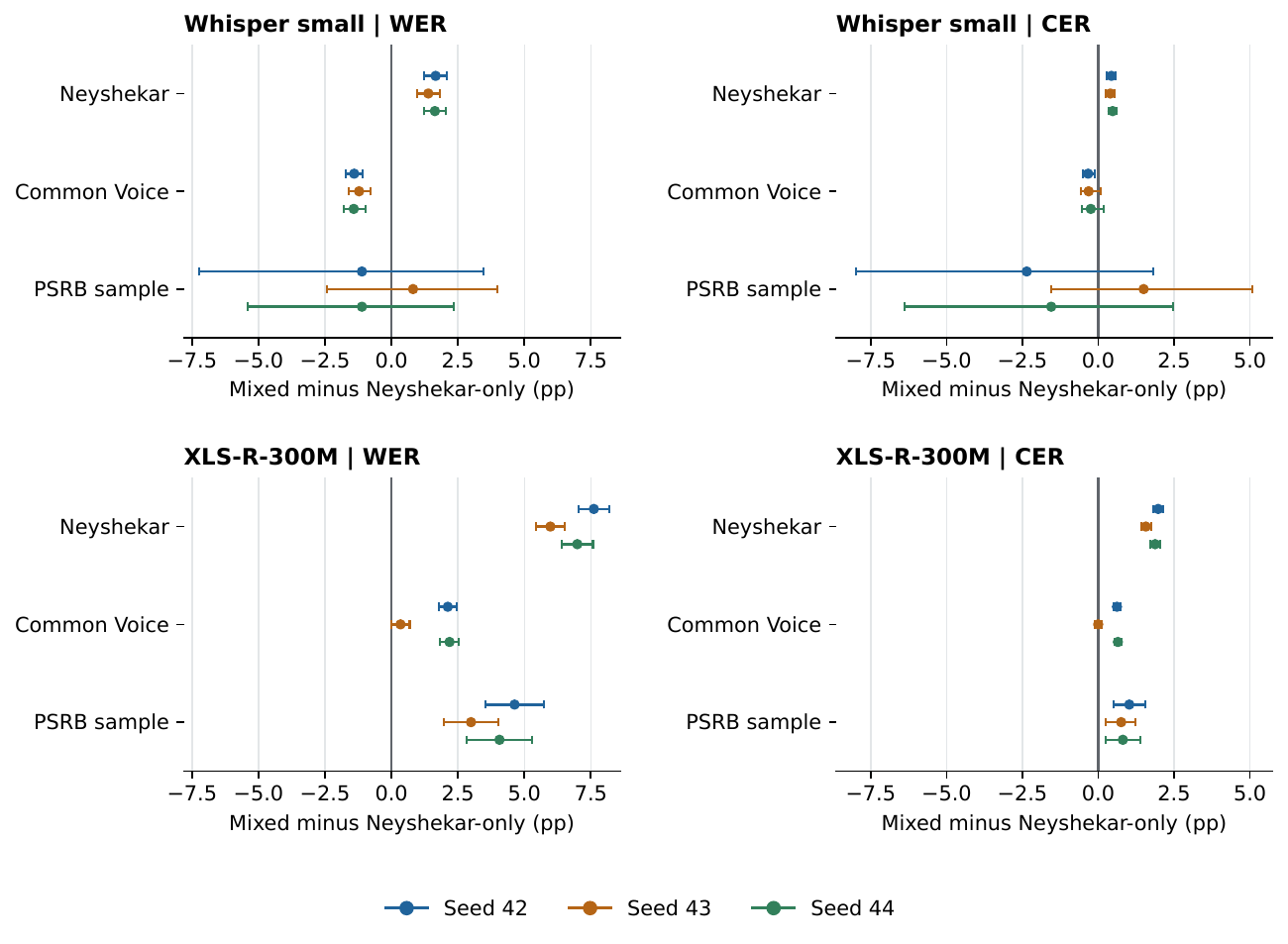}
\caption{The 64h mixture control at 2,382 updates: mixed minus Neyshekar-only WER/CER in percentage points. Negative differences favour mixing. Each point and interval represents one optimisation seed and its paired percentile 95\% bootstrap interval. Utterances are resampled within each test. Intervals are conditional on trained systems and are not multiplicity-adjusted.}
\label{fig:mixture-results}
\end{figure}

\end{document}